\documentclass[11pt,a4paper]{article}
\usepackage[hyperref]{acl2020}
\usepackage{times}
\usepackage{latexsym}

\usepackage{comment}
\usepackage{subcaption}

\usepackage{arabtex}
\usepackage{utf8}

\usepackage{booktabs}
\usepackage{multirow}
\usepackage{graphicx}

\usepackage{microtype}

\aclfinalcopy 

\title{AraDIC: Arabic Document Classification using Image-Based Character Embeddings and Class-Balanced Loss}

\author{Mahmoud Daif, Shunsuke Kitada, Hitoshi Iyatomi\\
  Hosei University \\
  Graduate School of Science and Engineering \\
  Department of Applied Informatics \\
  \texttt{\{mahmoud.daif.8h@stu., shunsuke.kitada.8y@stu., iyatomi@\}} \\
  \texttt{ hosei.ac.jp}}
\date{}
\begin{document}
\maketitle
\begin{abstract}
\vspace*{-0.6cm}

    Classical and some deep learning techniques for Arabic text classification often depend on complex morphological analysis, word segmentation, and hand-crafted feature engineering.
These could be eliminated by using character-level features.
We propose a novel end-to-end Arabic document classification framework, Arabic document image-based classifier (AraDIC), inspired by the work on image-based character embeddings.
AraDIC consists of an image-based character encoder and a classifier.
They are trained in an end-to-end fashion using the class balanced loss to deal with the long-tailed data distribution problem.
To evaluate the effectiveness of AraDIC, we created and published two datasets, the Arabic Wikipedia title (AWT) dataset and the Arabic poetry (AraP) dataset.
To the best of our knowledge, this is the first image-based character embedding framework addressing the problem of Arabic text classification. 
We also present the first deep learning-based text classifier widely evaluated on modern standard Arabic, colloquial Arabic and classical Arabic.
AraDIC shows performance improvement over classical and deep learning baselines  by 12.29\% and 23.05\% for the micro and macro F-score, respectively.
\end{abstract}

\section{Introduction}
    \setcode{utf8}
Arabic is one of the six official languages of the United Nations and the official language of 26 states.
It is spoken by as many as 420 million people making it the fifth most popular language worldwide.
According to the Internet World Statistics, as of 2017, Arab users represent 4.8\% of internet users\footnote{Arabic Speaking Internet Users and Population Statistics. \url{https://www.internet-worldstats.com/stats19.html} Accessed: 16-Dec-2018,}.

Arabic can be classified into three different types each having its own purpose and morphology.
The modern standard Arabic, the colloquial or dialectal Arabic and the classical or old Arabic. 
The modern standard Arabic is the official language used in media, government, news papers and is taught in schools. 
Colloquial Arabic varies between countries and regions. 
Old or classical Arabic survives nowadays in religious scriptures and old poetry.

Arabic has 28 basic letters all are consonants except three, which are long vowels.
Arabic is written from right to left.
Most Arabic letters have more than one written form depending on their position in the word.
For example,
``~\<س>~''.
``~\<ـس>~'',
``~\<ـسـ>~'', and 
``~\<سـ>~'' 
are all different forms of the letter
``~\<س>~''(sīn).
In addition, diacritical marks/short vowels that contribute to the phonology of Arabic, greatly alter the character shape. 
Example, 
``~\<بً>~'',
``~\<بٌ>~'',
``~\<بٍ>~'',
``~\<بّ>~'',
``~\<بْ>~'',
``~\<بُ>~'',
``~\<بَ>~'', and 
``~\<بِ>~''
are combination of the letter 
``~\<ب>~''(bā') 
with different diacritics.
This visual nature of the Arabic letters is the main motivation for us to use image based embeddings.

The importance of text classification has increased due to the increase of textual data on the internet as a result of social networks and news sites.
Common examples of text classification are sentiment analysis~\cite{ibrahim2015sentiment}, spam detection~\cite{el2009filtering} and news categorization~\cite{shehab2016supervised}.
Arabic text classification is particularly challenging because of its complex morphological analysis.

Most research on Arabic text classification has used classical techniques for feature extraction~\cite{salloum2018survey}, which require complex morphological analysis, such as negation handling\cite{al2016arasenti}, part of speech tagging~\cite{khoja2001apt}, stemming~\cite{al2015novel}, and segmentation~\cite{abdelali2016farasa}. 
Arabic segmentation is especially complex because Arabic words are not always separated by white spaces.
It also includes some hand-crafted features like document term matrix with term frequency inverse document frequency (TF-IDF) scores or word count. 

Arabic text classification have been often done using classical algorithms like support vector machines (SVMs) or Naive Bayes~\cite{salloum2018survey}. 
Despite advances of text classification using deep learning techniques, little work has been done on Arabic. 
\citet{soliman2017aravec} introduced AraVec, which is a pretrained distributed word embeddings~\cite{mikolov2013distributed}. 
They trained their model using the skip-gram and continuous bag of words techniques.
They used data from different sources like Wikipedia and Twitter. 
More recently, \citet{sagheer2018arabic} used AraVec's pretrained word embeddings with sentence convolutional neural network (CNN) originally proposed by \citet{kim2014convolutional} for Arabic document classification. 
This method still did not mitigate the problem of Arabic word segmentation. 

Those combinations left two major issues unaddressed. 
First, performance highly depends on morphological analysis and word segmentation, which is difficult for Arabic. 
The same problem has been addressed for languages such as Japanese and Chinese~\cite{peng2003text}.
Second, obtaining appropriate embedding (i.e. building hand-crafted features) is difficult.

To solve these problems, character-based approaches utilizing deep learning methods mainly used in image processing have been proposed \cite{zhang2015character, shimada2016document, kitada2018end}.

\citet{zhang2015character} introduced a character-level CNN (CLCNN) that treats text as a raw signal at character level. 
The CNN then learns the language morphology and extracts appropriate features for text classification. 
Their method mitigated the issue of complex morphological analysis.

After that, \citet{shimada2016document} proposed image-based character embeddings for Japanese and Chinese text classification. 
Their model was composed of a convolutional auto-encoder (CAE)~\cite{masci2011stacked} and a CLCNN.
They were the first to handle a character as an image and obtained character-embedding with their CAE. 
They also introduced wild card training as a data augmentation technique, which is dropout~\cite{srivastava2014dropout} on the embedding space. 

Later, \citet{liu2017learning} used image-based character embeddings learned through a character encoder (CE) to train a gated recurrent unit (GRU) for Japanese, Chinese, and Korean text classification.

\citet{kitada2018end} proposed CE-CLCNN that concatenated \citet{liu2017learning}'s CE with CLCNN as an end-to-end system and introduced random erasing on image domain as a data augmentation method.
These models using character-level features learn language morphology eliminating the need for complex morphological analysis and word segmentation.

Another problem is that large text classification datasets usually suffer from long tailed data distribution problem.
This means that few classes make up majority of data.
This problem often reduces the model's accuracy on the minority classes making more biased towards majority classes.

This problem can be addressed by either re-sampling~\cite{chawla2002smote,shen2016relay, geifman2017deep, buda2018systematic, zou2018domain} or re-weighting the cost function~\cite{ting2000comparative, zhou2005training, huang2016learning, khan2017cost, cui2019class}.

\citet{cui2019class} noticed that re-weighting the cost function by inverse class frequency as used in vanilla schemes \cite{huang2016learning, huang2019deep,wang2017learning} could lead to poor performance on majority classes. They proposed class-balanced (CB) loss based on the effective number of classes which re-weights the loss by the inverse of the effective number of classes. 

Our contributions can be summarized as follows: 
\begin{itemize}
    \item We propose AraDIC which is a framework for Arabic text classification. AraDIC is an end-to-end model of a character encoder and a classifier trained using CB loss. 
    \item CB loss was originally intended for object detection tasks. We show that it can solve class-imbalance problems for text classification tasks.
    \item We introduce two datasets in the hope of becoming bench marking datasets for Arabic text classification tasks as well. 
    The Arabic Wikipedia title (AWT) dataset and the the Arabic poetry (AraP) dataset. 
    These two datasets contain the three types of Arabic language. 
\end{itemize}
To the best of our knowledge, this is the first time an image-based character embedding model is used for Arabic text classification.
Also, the first time a deep-learning based model is tested on datasets containing the three types of Arabic. 
This shows that our method could be used to overcome Arabic's complicated morphological analysis and word segmentation for all types of Arabic.
The code and datasets are released at \url{https://github.com/mahmouddaif/AraDIC}
    
\section{Datasets}
    \begin{figure}[t]
    \centering
    \begin{minipage}{\linewidth}
        \includegraphics[width=\linewidth]{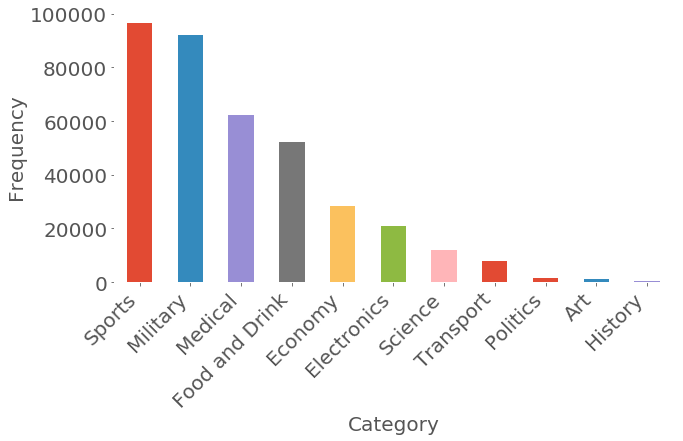}
        \subcaption{}
        \label{WikipediaTitleCatDist}
    \end{minipage} \\
    \begin{minipage}{\linewidth}
        \includegraphics[width=\linewidth]{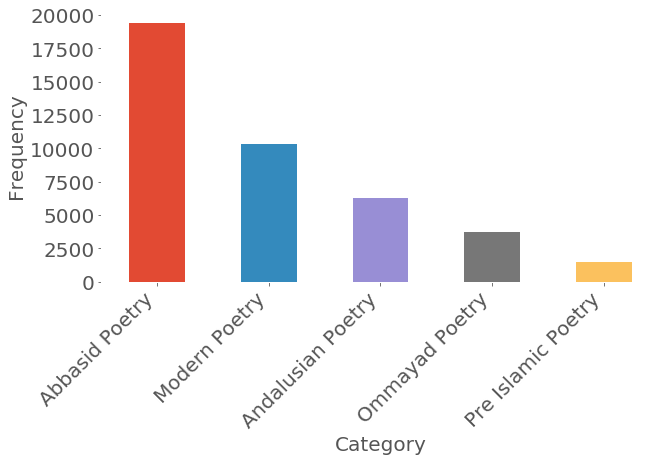}
        \subcaption{}
        \label{PoetryCatDist}
    \end{minipage}
    \caption{The category distribution for the (a) AWT and (b) AraP datasets.}
    \label{fig:dataset_distribution}
\end{figure}
\begin{table}[t]
    \centering
    \begin{minipage}{\linewidth}
        \begin{tabular}{@{}cc@{}}\\
            \toprule
            Layer & Configuration \\ \midrule
            Conv2D & (c= 1, k = 3x3, f=32) + ReLU \\
            Max-Pool2D & (k=2x2) \\
            Conv2D & (c=32, k = 3x3, f=32) + ReLU \\
            Max-Pool2D & (k=2x2) \\
            Conv2D & (c=32, k = 3x3, f=32) + ReLU \\
            FC & (800,128) + ReLU \\
            FC & (128,128) + ReLU \\ \bottomrule
        \end{tabular}
        \subcaption{Character encoder architecture.}
        \label{tab:CEArchTable}    
    \end{minipage} \\
    \begin{minipage}{\linewidth}
    \end{minipage} \\
    \begin{minipage}{\linewidth}
        \begin{tabular}{@{}cc@{}}
            \toprule
            Layer & Configuration \\ \midrule
            Conv1D & (c= 128, k = 3, f=512) + ReLU \\
            Max-Pool1D & (k=3) \\
            Conv1D & (c=512, k=3, f=512) + ReLU \\
            Max-Pool1D & (k=3) \\
            Conv1D & (c=512, k = 3, f=512) + ReLU \\
            Conv1D & (c=512, k = 3, f=512) + ReLU \\
            FC & (1024,1024) + ReLU \\
            FC & (1024,nc) + ReLU \\ \bottomrule
        \end{tabular}
        \subcaption{CLCNN architecture.}    
        \label{tab:CLCNNArchTable}
    \end{minipage}
    \begin{minipage}{\linewidth}
    \end{minipage}
    \begin{minipage}{\linewidth}
        \begin{tabular}{@{}cc@{}}
            \toprule
            Layer & Configuration \\ \midrule
            BiGRU & (input = 128, hidden = 128, layer = 3) \\
                &  + BN \\
            FC & (256,nc) \\ \bottomrule
            \end{tabular}
        \subcaption{BiGRU architecture.}
        \label{tab:BiGRUArchTable}
    \end{minipage}

    \caption{AraDIC's architectural configuration, \textbf{c} is input channels, \textbf{k} is kernel size, \textbf{f} is feature maps, \textbf{nc} is number of classes and \textbf{BN} is Batch Normalization~\cite{ioffe2015batch}.}
\end{table}
\begin{figure}[t]
    \centering
    \includegraphics[width=\linewidth]{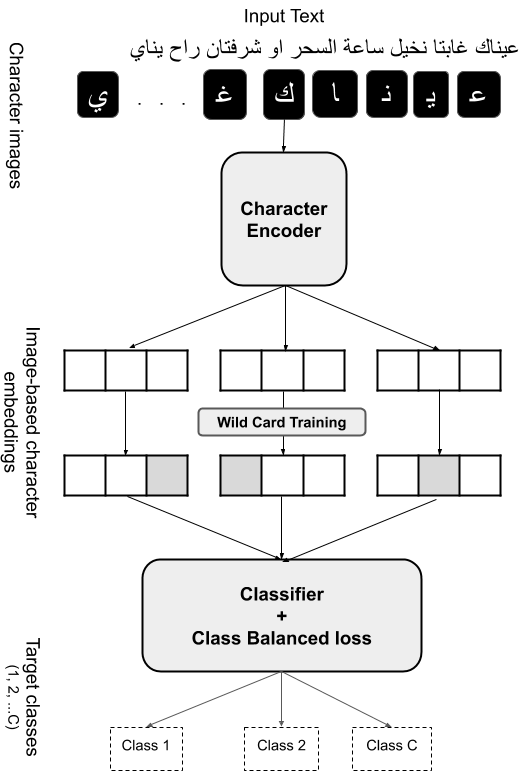}
    \caption{
        AraDIC's architecture outline.
    }
    \label{CECLCNNfig}
\end{figure}

Arabic text classification lacks bench marking datasets. 
This is because it is expensive and time consuming to annotate a large dataset to be used for text classification using deep learning algorithms. 
We created two large datasets that do not require manual annotation and can be used as benchmarks for Arabic text classification. 
The AWT and the AraP datasets.
Sections \ref{AWTD} and \ref{APD} describe how we constructed these datasets.

\subsection{ Arabic Wikipedia Title Dataset (AWT) } \label{AWTD}
\citet{liu2017learning} introduced the Wikipedia title dataset for Japanese, Chinese and Korean by making use of Wikipedia's recursive hierarchical structure to crawl 12 different Wikipedia categories and using the category as a label to all article titles under this category, and its subcategories.
He assumed that an article only exists in one category. 
If an article existed in more that one category, it was randomly assigned to only one of them. 
This created some noisy annotations, however, categories were chosen as distinctive in nature as possible to reduce this problem.
We crawl 11 different categories from the Arabic Wikipedia using the same method. 
A total of 444,911 different titles with a total of 4,196,127 different words were crawled.
This dataset contains mostly modern standard Arabic.
The dataset category distribution can be found in Figure~\ref{WikipediaTitleCatDist}.

\subsection{ Arabic Poetry Dataset (AraP) }\label{APD}
The AraP dataset was crawled from the Adab Website\footnote{Adab website for Arabic poetry from 6th to 21st centuries. \url{http://www.adab.com/}.} 
It contains Arabic poetry from the 6th to 21st centuries and consists of 41,264 poems from five eras.
This dataset contains mostly colloquial and old Arabic.
AraP's Category distribution details can be found in Figure~\ref{PoetryCatDist}.
\begin{table*}[t]
\centering
\resizebox{\textwidth}{!}{
\begin{tabular}{@{}lllrrrr@{}}
\toprule
\multicolumn{3}{c}{\multirow{2}{*}{Model}} & \multicolumn{4}{c}{F-score}                                                                                                       \\ \cmidrule(l){4-7} 
\multicolumn{3}{c}{}                       & \multicolumn{2}{c}{Arabic Wikipedia Title}                      & \multicolumn{2}{c}{Arabic Poetry}                               \\ \cmidrule(lr){4-5} \cmidrule(lr){6-7}
           & Embedding & Classifier        & \multicolumn{1}{c}{Micro [\%]} & \multicolumn{1}{c}{Macro [\%]} & \multicolumn{1}{c}{Micro [\%]} & \multicolumn{1}{c}{Macro [\%]} \\ \cmidrule(r){1-1} \cmidrule(lr){2-2} \cmidrule(lr){3-3} \cmidrule(lr){4-4} \cmidrule(lr){5-5} \cmidrule(lr){6-6} \cmidrule(lr){7-7}
       &    &         Majority Class       & 21.67                          & 2.97                          & 47.06                          & 5.33                          \\ \cmidrule(r){1-1} \cmidrule(lr){2-2} \cmidrule(lr){3-3} \cmidrule(lr){4-4} \cmidrule(lr){5-5} \cmidrule(lr){6-6}  \cmidrule(lr){7-7}
Word       & Unigram   & SVM               & 45.47                          & 26.60                          & 52.80                          & 34.83                          \\
level      & AraVec    & CNN               & 45.02                          & 25.05                          & 69.28                          & 41.95                          \\ \cmidrule(r){1-1} \cmidrule(lr){2-2} \cmidrule(lr){3-3} \cmidrule(lr){4-4} \cmidrule(lr){5-5} \cmidrule(lr){6-6}  \cmidrule(lr){7-7}
Character  & One-hot   & CLCNN             & 42.76                          & 18.71                          & 68.24                          & 37.72                          \\
level      & \textbf{AraDIC}     & CLCNN ($-$ CB loss) & 47.47                          & 26.85                          & 74.86                          & 45.61                          \\
           &           & CLCNN ($+$ CB loss) & 49.49                          & 30.55                          & 74.03                          & 48.65                          \\ 
           &           & BiGRU ($-$ CB loss)   & 55.71         &              39.04                  &              78.93          &                59.88             \\
           &           & BiGRU ($+$ CB loss)   &   \textbf{57.76}                 &    \textbf{44.54}                 &             \textbf{79.53}           &                   \textbf{65.00}             \\\bottomrule

\end{tabular}}
\caption{Classification results of our model and other baselines. \textbf{Majority Class}: Due to high class-imbalance in both of our datasets, we examine the performance of majority class classifier. \textbf{CNN + AraVec}: Sentence classifier CNN~\cite{sagheer2018arabic, kim2014convolutional} using AraVec's word embeddings~\cite{soliman2017aravec}. \textbf{SVM}: an SVM with unigrams, stemming, and document term matrix with TF-IDF scores as features. \textbf{CLCNN}: character level CNN with one hot encoding as inputs\cite{zhang2015character}. \textbf{AraDIC}: our proposed end-to-end framework of character encoder, CLCNN and BiGRU classifiers, trained with and without class-balanced softmax loss \textbf{(CB loss)}. We report two evaluation metrics, the macro and micro F-scores.}
\label{tab:result}
\end{table*}

\section{Methodology}
    
AraDIC is an end-to-end framework of a character encoder (CE) and a classifier. We choose two classifiers for our framework. A character CNN (CLCNN) similar to \citet{kitada2018end}, but tuned to Arabic language, and a bidirectional gated recurrent unit (BiGRU)~\cite{chung2014empirical} based classifier. 
The outline of our framework is shown in Figure~\ref{CECLCNNfig}. 
We use wildcard training introduced by \citep{shimada2016document} for data augmentation.
Wildcard training is dropout on the embedding space so that the data changes a little every training iteration. In that sense it acts as a data augmentation technique.
We use CB softmax loss to deal with class imbalance problem.
\subsection{Character Encoder}\label{CE}
The CE is a CNN where convolution is performed in a depth-wise manner.
It learns to encode each input character image of size 36 $\times$ 36 pixels into a 128-dimension vector. The architectural configuration is shown in Table~\ref{tab:CEArchTable}.

\subsection{Classifier}\label{CLCNNSS}
For classification we use two classifiers. 
The first one is a CLCNN, and the second is a BiGRU. 
Input text is represented as an array of character images each encoded into a 128 dimension vector using the CE.
Those character embeddings are the input features for both the CLCNN and the BiGRU.

The CLCNN is a character-level CNN whose architectural details can be found in Table~\ref{tab:CLCNNArchTable}.

The BiGRU takes those characters embeddings and computes a sentence level embedding. The sentence embedding is the average of all the hidden layers outputs of the BiGRU. 
These sentence level features are then passed to a fully connected layer followed by a softmax for class prediction.
Detailed architecture of the BiGRU can be found in Table~\ref{tab:BiGRUArchTable}.

\subsection{Class-Balanced Loss}\label{BalancedLoss}
Both of our datasets suffer from the long tailed distribution problem as shown in Figure~\ref{WikipediaTitleCatDist} and \ref{PoetryCatDist}. 
To deal with this problem, we use state-of-the-art method, the class balanced loss~\cite{cui2019class}. 
The class-balanced loss could be applied by re-weighting the loss function by the inverse effective number of classes. We apply it to softmax cross entropy loss as follows:
\begin{equation} \label{eq:classbalancedsoftmax}
- \frac{1-\beta}{1-\beta^{n_{y}}} \log\left(\frac{\exp{(Z_{y})}}{\sum_{j=1}^{C} \exp{(Z_{j})}}\right),
\end{equation}
where $\frac{1-\beta}{1-\beta^{n_{y}}}$ is the inverse effective number of classes.  $Z_{j}$ is the model output $(j = 1, 2, ... C)$, $y$ is class label for the input sample, $n_{y}$ is number of samples per class $y$ and $\beta$ is a training hyper parameter.
This will assign adaptive weights to the cost function for classes with higher samples and classes with lower samples, effectively re-weighting the cost function based on effective number of classes. 
This method was originally intended for object detection, we show that it can be applied to text classification as well. 
\begin{figure*}[t]
    \centering
    \includegraphics[width=\linewidth]{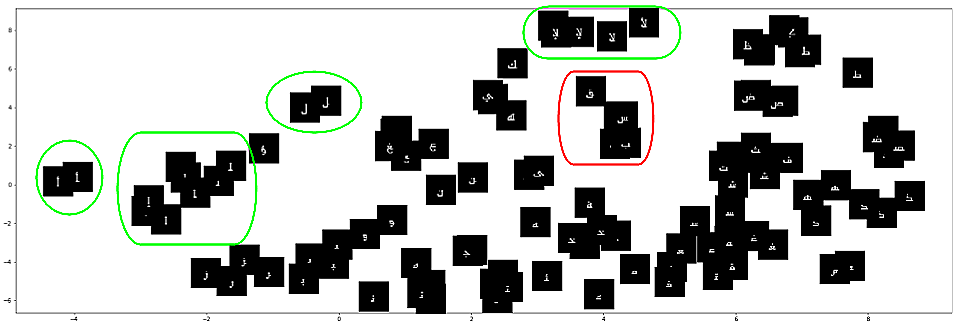}
    \caption{
        Character embeddings visualization using t-SNE~\cite{maaten2008visualizing}. Sections circled in green show clusters of related characters with similar shapes, which was the majority of cases. Sections encircled in red show clusters of unrelated characters which was rare.
    }
    \label{charembeddings}
\end{figure*}

\section{Experiments}
    
To train our classifier both datasets are divided into 80\% training data and 20\% testing data\footnote{Hyperparameters were tuned with a validation set split from the training set, and reported the predicted results of the evaluation set.}.

\subsection{AraDIC}
The maximum character length or each document is set to 60 characters for the AWT dataset and 128 characters for the AraP dataset. 
That's for using the CLCNN classifiers. 
As for the BiGRU classifier we don't set a maximum character length, instead the whole text is used.
Each character was encoded into a 128 dimension vector using the CE. 
Adam optimizer~\cite{kingma2014adam} with a batch size of 64 and a learning rate of 0.001 was chosen as the optimization method. 
As for the CB loss we set $\beta$ to 0.99 for both datasets.
Wildcard training ratio is set to 10\%.
The training loss converged after approximately 150 epochs for AraP dataset and 500 epochs for AWT dataset. 

\subsection{Baselines}
We use several word-based and character-based baselines to evaluate our method. They include both classical and deep learning baselines as follows:
\begin{itemize}
 \item Due to high class imbalance in both our datasets,  a majority class classifier is chosen as our first baseline.
  \item A classical Support Vector Machine (SVM) with a document-term matrix (DTM) of TF-IDF scores for unigrams as input was used as word-based baseline. Terms occurring only once and terms appearing in more than 90\% of documents were omitted from the DTM. We performed preprocessing in the form of stop words, non-Arabic characters, diacritics removal. Then, text is stemmed using Khoja stemmer~\cite{khoja2001apt}. Farasa segmenter~\cite{abdelali2016farasa} was used for word segmentation.
  \item We also used \citet{sagheer2018arabic}'s method of using AraVec's word embeddings as input features and sentence CNN originally introduced by \citet{kim2014convolutional} for classification. This is another word-based baseline.
  \item Another baseline is a character-level CNN (CLCNN) introduced by~\citet{zhang2015character}. In this baseline, input characters were one-hot encoded.
\end{itemize}
. 

\section{Results and Discussion}
\setcode{utf8}

Classification results can be found in Table~\ref{tab:result}.
It is noted that AraDIC outperforms both word based and character based deep learning and classical baselines.
Performance improvement is shown over classical SVM without the need for preprocessing, word segmentation, stemming and feature engineering associated with classical methods. 
It was also able to beat \citet{sagheer2018arabic} method of using sentence CNN with AraVec's word embeddings as input features without the need for word segmentation.
This makes character level representations a better choice for Arabic language avoiding segmentation and feature engineering problems.
It's also shown that using AraDIC's image-based character embeddings outperforms CLCNN with one-hot encoded characters as input features.
Therefore, we can conclude as well that image-based character embeddings are useful for Arabic language due to the property of the language as discussed in the introduction section of this paper.

As for the classifier part of AraDIC, it can be noticed that the BiGRU significantly outperforms CLCNN for both classification tasks. 
This suggests that sequence-to-sequence models are more suitable for text classification using image-based character-based embeddings, especially in Arabic document classification.

Also, using CB loss improves the macro F-score of classifiers for both datasets.
It can be also noted that the improvement in the macro F-score is achieved when using a CLCNN and a BiGRU.
This shows that CB loss can be useful to solve class imbalance problems for text classification tasks.

Figure~\ref{charembeddings} shows character embeddings visualization using t-distributed stochastic neighbor embedding (t-SNE) method~\cite{maaten2008visualizing}.
As shown, embedding for related characters having similar shapes like
``~\<أ>~'',
``~\<ا>~'',
``~\<ا>~'', and 
``~\<إ>~''
are clustered in the embedding space.
This is the majority of cases. 
Other unrelated characters are also clustered which is rare. 
This however shows that using image based character embeddings gives an extra layer of visual information.
Another reason why it is useful is because both the CE and the classifier are trained as an end-to-end system.
This means that the CE learns the best embeddings suitable for the classifier.

\section{Conclusion}
In this paper, we proposed a novel end-to-end Arabic text classification framework AraDIC.
We also published two large scale Arabic text classification datasets that contain the three types of Arabic language, the AWT and the AraP datasets.
AraDIC’s image-based character embedding strategy eliminated
the need for complicated preprocessing, segmentation and
morphological analysis, and achieved much better performance than
conventional deep and classical text classification techniques that use word and character-based embeddings. 
We have shown also that class-balanced loss is useful for text classification tasks with long tailed distribution datasets.

\bibliography{acl2020}
\bibliographystyle{acl_natbib}

\end{document}